\documentclass[11pt,a4paper]{article}

\usepackage[margin=1in]{geometry} 
\usepackage{times} 

\usepackage{amsmath}
\usepackage{amssymb}
\usepackage{mathtools}

\usepackage{graphicx}
\usepackage{booktabs}
\usepackage{multirow}
\usepackage{adjustbox}
\usepackage{threeparttable}
\usepackage[figuresright]{rotating}
\usepackage{siunitx}
\usepackage{tabularx}
\usepackage{placeins} 

\usepackage{algorithm}
\usepackage{algpseudocode}

\usepackage{textcomp}
\usepackage{xcolor}
\usepackage{url}
\usepackage[bookmarks=false, colorlinks=true, linkcolor=blue, citecolor=blue, urlcolor=blue]{hyperref}
\usepackage[numbers, sort&compress]{natbib} 
\usepackage{authblk} 

\usepackage{tikz}
\usetikzlibrary{shapes,shapes.geometric, arrows, positioning, calc,arrows.meta, quotes}
\usepackage{tcolorbox}

\newcommand{\modelname}{\textsc{CGPT}}
\newcommand{\vect}[1]{\mathbf{#1}}
\newcommand{\mat}[1]{\mathbf{#1}}

\tikzset{
 block/.style = {rectangle, draw, fill=blue!10, text width=6em, text centered, rounded corners, minimum height=4em},
 process/.style = {rectangle, rounded corners, minimum width=3cm, minimum height=1cm, text centered, draw=black, fill=orange!30},
 io/.style = {trapezium, trapezium left angle=70, trapezium right angle=110, minimum width=3cm, minimum height=1cm, text centered, draw=black, fill=blue!30},
 arrow/.style = {thick,->,>=stealth}
}

\title{\textbf{Causally-Guided Pairwise Transformer - Towards Foundational Digital Twins in Process Industry}}

\author[1]{Michael Mayr\thanks{Corresponding author: michael.mayr@scch.at}}
\author[1]{Georgios C. Chasparis}

\affil[1]{Software Competence Center Hagenberg, Softwarepark 32a, 4232 Hagenberg im Mühlkreis, Austria}

\date{} 

\begin{document}
\maketitle
\flushbottom

\begin{abstract}
\noindent Foundational modelling of multi-dimensional time-series data in industrial systems presents a central trade-off: channel-dependent (CD) models capture specific cross-variable dynamics but lack robustness and adaptability as model layers are commonly bound to the data dimensionality of the tackled use-case, while channel-independent (CI) models offer generality at the cost of modelling the explicit interactions crucial for system-level predictive regression tasks. To resolve this, we propose the Causally-Guided Pairwise Transformer (CGPT), a novel architecture that integrates a known causal graph as an inductive bias. The core of CGPT is built around a pairwise modeling paradigm, tackling the CD/CI conflict by decomposing the multidimensional data into pairs. The model uses channel-agnostic learnable layers where all parameter dimensions are independent of the number of variables. CGPT enforces a CD information flow at the pair-level and CI-like generalization across pairs. This approach disentangles complex system dynamics and results in a highly flexible architecture that ensures scalability and any-variate adaptability.
We validate CGPT on a suite of synthetic and real-world industrial datasets on long-term and one-step forecasting tasks designed to simulate common industrial complexities. Results demonstrate that CGPT significantly outperforms both CI and CD baselines in predictive accuracy and shows competitive performance with end-to-end trained CD models while remaining agnostic to the problem dimensionality.

\vspace{1em}
\end{abstract}

\section{Introduction}
\label{sec:introduction}

\noindent The European process industry faces increasing pressures from economic competition and regulatory demands, particularly concerning energy efficiency and greenhouse gas emission reduction targets (e.g., EU 2050 \cite{EU2050}). To maintain global competitiveness, staying on top of industrial and scientific advancements is a necessity. The growth of retrofitted sensors across various sectors has led to an explosion in data volume \cite{Lasi2014}, offering opportunities to leverage complex information for enhanced operational efficiency and decision-making. Digital Twins (DTs) have emerged as a key enabling technology in this context \cite{Kritzinger2018, Rasheed2020}. Recent advancements highlight a progression towards more autonomous and intelligent systems, leading to the concept of \emph{Cognitive Digital Twins} (see  \cite{Abburu2020, Mayr2025}). A cognitive DT should be capable of robustly learning from and providing highly autonomous decision support for diverse tasks and use-cases across a wide range of multi-dimensional industrial datasets. \emph{Transfer Learning (TL)} and \emph{Self-Supervised Learning (SSL)} paradigms are considered a potential enabler for developing these advanced DTs and to move from task-specific implementations to foundational DTs \cite{Mayr2024review}.

\subsection{The Challenge: Towards Generalist Industrial Foundation Models}
\noindent The ambitious goal for industrial foundational DTs is to engineer a single, pre-trained foundation model capable of generalizing across an entire corporation's diverse set of use-cases. A primary obstacle to achieving this goal lies in addressing the \textit{any-variate} problem and ensuring \textit{cross-domain} adaptability. Industrial environments are commonly dynamic; sensors are frequently added, removed, or upgraded, which constantly changes the shape of the input data. Furthermore, industrial use-cases are diverse and heterogeneous, e.g. predicting different Key Performance Indicators (KPIs) often requires modelling only a unique subset of the available features for each specific task. Thus, a truly generalist model must be capable of handling highly heterogeneous and any-variate predictive tasks. This means the model needs to operate seamlessly, regardless of the number or type of input variables it encounters.

\subsection{Multidimensional Forecasting for Digital Twin Use-Cases}
\label{sec:problem_formulation}
\noindent Industrial Digital Twin (DT) systems often exhibit complex dynamics, including lagged effects and slow inertia, where changes in control inputs might not immediately affect Key Performance Indicators (KPIs). Industrial time-series modelling faces challenges in capturing highly lagged and non-linear causal relationships without prohibitive computational costs. Deep learning models, like our \modelname, generally address the complexity by formulating the problem as \emph{historic context to horizon prediction}, focusing on modelling the influence of specific historical "context" variables on future "target" variables within defined windows. This approach avoids fitting a global, high-order dynamic model for all variables simultaneously, enhancing computational efficiency and generalizability.
This paper focuses on time series forecasting and broader \emph{predictive regression} tasks where the output has some autoregressive dependencies on its own history. For instance, predicting a KPI ($\mathbf{Y}_{t+h}$) may depend on its own history, other process variables, and control inputs within $\mathcal{S}_{hist}$. 

\subsection{Limitations of Existing Time-Series modelling Paradigms}
\noindent \emph{Channel-Independent (CI) models}, such as PatchTST \cite{Nie2023patchtst} or DLinear \cite{Zeng2023dlinear}, process each sensor stream in isolation (see (a) CI Model in \ref{fig:cdvci}). They achieve remarkable robustness to noise and distribution shifts and often exhibit strong generalization capabilities, sometimes outperforming more complex Channel-Dependent (CD) approaches \cite{Zhao2024}. This success largely stems from their ability to decouple the learning problem, effectively achieving high transferability by treating each variable as an independent time series. However, their core assumption of independence is a fundamental limitation in physically interconnected systems. They are structurally blind to explicit physical cause-and-effect relationships governing industrial processes, which severely limits their utility for true system understanding and critical control applications like Model Predictive Control (MPC) or Reinforcement Learning (RL). \\
Conversely, \emph{Channel-Dependent (CD)} models, like a standard MLP, or the transformer-based  Crossformer \cite{Zhang2023crossformer}, jointly process all variables to capture their interplay (see (b) CD Model in \ref{fig:cdvci}). While designed for capturing multivariate dynamics, they are brittle in high-dimensional industrial environments. In such settings, a target variable's dynamics are often dictated by a sparse subset of potential causes, not all available sensors. These models are susceptible to "negative transfer" and their fixed input-output shape makes them difficult to adapt to new variates without retraining. Even minor changes in sensor configurations or task specification may necessitate complete retraining, undermining the concept of a truly generalist foundation model. \\

\tikzset{
    block/.style = {draw, fill=green!20, rectangle, minimum height=4.5em, minimum width=4.5em},
    arrow/.style={-Stealth, thick}, 
}
\tikzset{every edge quotes/.style =
    { fill = green!20,
      sloped,
      execute at begin node = $,
      execute at end node   = $  }}

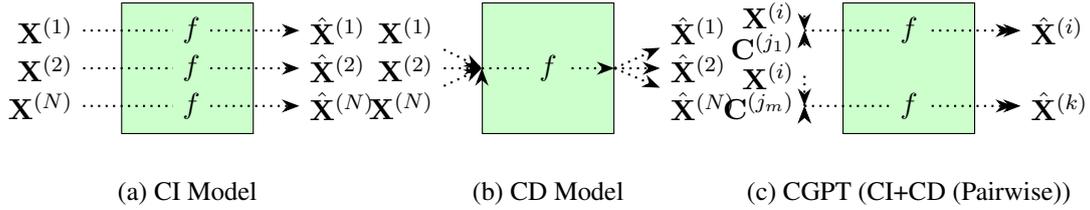
\begin{figure}[htp!]
\centering
\begin{tikzpicture}[node distance=2cm, >=Stealth]

    \node [block] (forecaster1) {};
    \node (input1) [left=0.5cm of forecaster1, yshift=0.5cm] {$\mathbf{X}^{(1)}$};
    \node (input2) [left=0.5cm of forecaster1] {$\mathbf{X}^{(2)}$};
    \node (input3) [left=0.5cm of forecaster1, yshift=-0.5cm] {$\mathbf{X}^{(N)}$};
    \node (output1) [right=0.6cm of forecaster1, yshift=0.5cm] {$\hat{\mathbf{X}}^{(1)}$};
    \node (output2) [right=0.6cm of forecaster1] {$\hat{\mathbf{X}}^{(2)}$};
    \node (output3) [right=0.6cm of forecaster1, yshift=-0.5cm] {$\hat{\mathbf{X}}^{(N)}$};
    
    \draw [arrow, dotted] (input1) edge ["f"] (output1);
    \draw [arrow, dotted] (input2) edge ["f"] (output2);
    \draw [arrow, dotted] (input3) edge ["f"] (output3);
    \node at (forecaster1.south) [below=0.5cm of forecaster1] {\small (a) CI Model};

    \node [block] (forecaster2) [right=3cm of forecaster1] {};
    \node (input2a) [left=0.5cm of forecaster2, yshift=0.5cm] {$\mathbf{X}^{(1)}$};
    \node (input2b) [left=0.5cm of forecaster2, yshift=0cm] {$\mathbf{X}^{(2)}$};
    \node (input2c) [left=0.5cm of forecaster2, yshift=-0.5cm] {$\mathbf{X}^{(N)}$};

    \node (output2a) [right=0.6cm of forecaster2, yshift=0.5cm] {$\hat{\mathbf{X}}^{(1)}$};
    \node (output2b) [right=0.6cm of forecaster2, yshift=0cm] {$\hat{\mathbf{X}}^{(2)}$};
    \node (output2c) [right=0.6cm of forecaster2, yshift=-0.5cm] {$\hat{\mathbf{X}}^{(N)}$};

    \draw [arrow, dotted] (input2a) -- (forecaster2.west);
    \draw [arrow, dotted] (input2b) -- (forecaster2.west);
    \draw [arrow, dotted] (input2c) -- (forecaster2.west);
    \draw [arrow, dotted] (forecaster2.west) edge ["f"'] (forecaster2.east); 
    \draw [arrow, dotted] (forecaster2.east) -- (output2a);
    \draw [arrow, dotted] (forecaster2.east) -- (output2b);
    \draw [arrow, dotted] (forecaster2.east) -- (output2c);
    \node at (forecaster2.south) [below=0.5cm of forecaster2] {\small (b) CD Model};

    \node [block] (forecaster3) [right=3cm of forecaster2] {};
    \node (input3a_target) [left=0.5cm of forecaster3, yshift=0.7cm] {$\mathbf{X}^{(i)}$};
    \node (input3a_context) [left=0.5cm of forecaster3, yshift=0.3cm] {$\mathbf{C}^{(j_1)}$}; 
    \node (input3b_target) [left=0.5cm of forecaster3, yshift=-0.1cm] {$\mathbf{X}^{(i)}$};
    \node (input3b_context) [left=0.5cm of forecaster3, yshift=-0.5cm] {$\mathbf{C}^{(j_m)}$}; 
    \node (output3a) [right=0.6cm of forecaster3, yshift=0.5cm] {$\hat{\mathbf{X}}^{(i)}$};
    \node (output3b) [right=0.6cm of forecaster3, yshift=-0.5cm] {$\hat{\mathbf{X}}^{(k)}$};
    
    \draw [arrow, dotted] (input3a_target.east) -- ([xshift=-0.5cm,yshift=0.5cm] forecaster3.west);
    \draw [arrow, dotted] (input3a_context.east) -- ([xshift=-0.5cm,yshift=0.5cm] forecaster3.west);
    \draw [arrow, dotted] ([xshift=-0.5cm,yshift=0.5cm]forecaster3.west) edge ["f"'] ([xshift=0.5cm,yshift=0.5cm]forecaster3.east); 
    \draw [arrow, dotted] ([xshift=0.5cm,yshift=0.5cm]forecaster3.east) -- (output3a);

    \draw [arrow, dotted] (input3b_target.east) -- ([xshift=-0.5cm,yshift=-0.5cm] forecaster3.west);
    \draw [arrow, dotted] (input3b_context.east) -- ([xshift=-0.5cm,yshift=-0.5cm] forecaster3.west);
    \draw [arrow, dotted] ([xshift=-0.5cm,yshift=-0.5cm]forecaster3.west) edge ["f"'] ([xshift=0.5cm,yshift=-0.5cm]forecaster3.east);
    \draw [arrow, dotted] ([xshift=0.5cm,yshift=-0.5cm]forecaster3.east) -- (output3b);
    
    \node at (forecaster3.south) [below=0.5cm of forecaster3] {\small (c) CGPT (CI+CD (Pairwise))};
\end{tikzpicture}
\caption{
\textbf{(a) Channel-Independent (CI):} A shared function $f$ processes each variable independently, learning universal temporal patterns but ignoring cross-channel dependencies.
\textbf{(b) Channel-Dependent (CD):} A single function $f$ processes all variables $(\mathbf{X}^{(1)}, \dots, \mathbf{X}^{(N)})$ jointly, i.e. mixed, to directly model their interdependencies.
\textbf{(c) \modelname\ (CI+CD Pairwise):} A single function $f$ processes pairs of  target\-context variables ($(\mathbf{X}^{(i)}$,$\mathbf{C}^{(j)})$) jointly (CD). All pairs w.r.t a specific target are processed independently (CI). The outputs from these pairwise computations are aggregated in the latent space (see Sec.~\ref{sec:cgpt}). This captures specific dependencies between the target and one context variable (CD) while generalizing across pairs (CI). (Figure adapted from \cite{Mayr2025}).
}
\label{fig:cdvci}
\end{figure}

\noindent While some recent approaches for foundational time-series models in the any-variate setting utilize S3, e.g \cite{Liu2024}, or a similar but more advanced concept is explored in Morai \cite{Woo2024}, where multivariate data are concatenated into a single long univariate stream, this can lead to excessively long context lengths and may still struggle to efficiently discern a variable's cause-effect relationships, a problem our pairwise approach directly addresses. TimesFM \cite{Das2023timesfm}, Tiny Time Mixer (TTM) \cite{Ekambaram2023} or TSMixer \cite{TSMixer} operate in a CI setting, with TTM introducing CI training and subsequent CD fine-tuning, freezing the pre-trained CI encoder. Those models have demonstrated impressive zero-shot forecasting capabilities, but the pre-trained part of the architecture primarily focuses on temporal patterns within uni-variate time-series.
To our knowledge, our work is the first that explores a pairwise modelling paradigm (see (c) CGPT (Pairwise CD modelling \ref{fig:cdvci}) using CI-like generalization while specifically modelling pairwise cause-effect relationships in the pre-training phase, distinguishing it from existing generalist models. Finally, \modelname~uses a pragmatic form of causal modelling, i.e. it avoids the difficult problem of causal discovery from observational time-series data; a causal graph is assumed; in industrial settings, such a causal graph may be constructed first in a data-driven way using Granger Causality or other methods; however, validation by domain experts is recommended. \modelname~focuses on the problem of leveraging these known cause-and-effect interactions to a target in both (pre-)training and inference.

\subsection{Contributions}
\label{sec:Contributions}
\noindent This paper proposes a \textbf{Causally-Guided Pairwise Transformer (CGPT)}, specifically tailored to the CI+CD training scheme (see Fig.~\ref{fig:cdvci}), and addresses the mentioned  challenges in foundational industrial modelling by providing a balanced approach between generality and specificity. The model operates on a fixed input size, i.e., two variates, enabling CI-like transfer to other datasets with a different number of variates, i.e., any-variate, a problem that is non-trivial for CD models when layers are not variate-agnostic. The architecture and methodology are described in Sec.~\ref{sec:cgpt}, where experiments with real-world and synthetic data show competitiveness with end-to-end trained CD time-series models, e.g. MLPs, which have fixed input dimensionalities. We also conduct experiments (see Sec.~\ref{sec:experiments}) against channel-independent univariate model (e.g. DLinear \cite{Zeng2023dlinear}) which commonly serves as strong baseline.

\section{\modelname: Causally-Guided Pairwise Transformer - Towards Foundational Digital Twins}
\label{sec:cgpt}

\noindent Our objective is to develop a function $f_\theta$ that maps the historical state of a multivariate industrial process to the future value of a specific target variable, effectively capturing intricate, causally-informed temporal dependencies. Let $\vect{C}_t = (C_1(t), \dots, C_{N_{vars}}(t)) \in \mathbb{R}^{N_{vars}}$ represent the multivariate state of the process at discrete time step $t$. The individual channels $C_n(t)$ in this state vector can be broadly categorized:

\begin{itemize}
    \item \textbf{Internal Process State Measurements}: Continuously sensed data that directly describe the physical or chemical state of the process (e.g., temperatures inside a reactor, pressures in a pipeline, motor current, air flow). These are typically reactive measurements reflecting the process's internal dynamics.
    \item \textbf{Operational and Product-Specific Variables}: Variables characterizing the operational configuration or external context. This includes actively controlled parameters (e.g., machine setpoints, valve positions) and variables defining the product or environment (e.g., product type, raw material properties, ambient humidity). These channels often act as causal drivers for the process behavior.
\end{itemize}

\begin{figure}[!ht]
    \centering
    \includegraphics[width=0.7\textwidth]{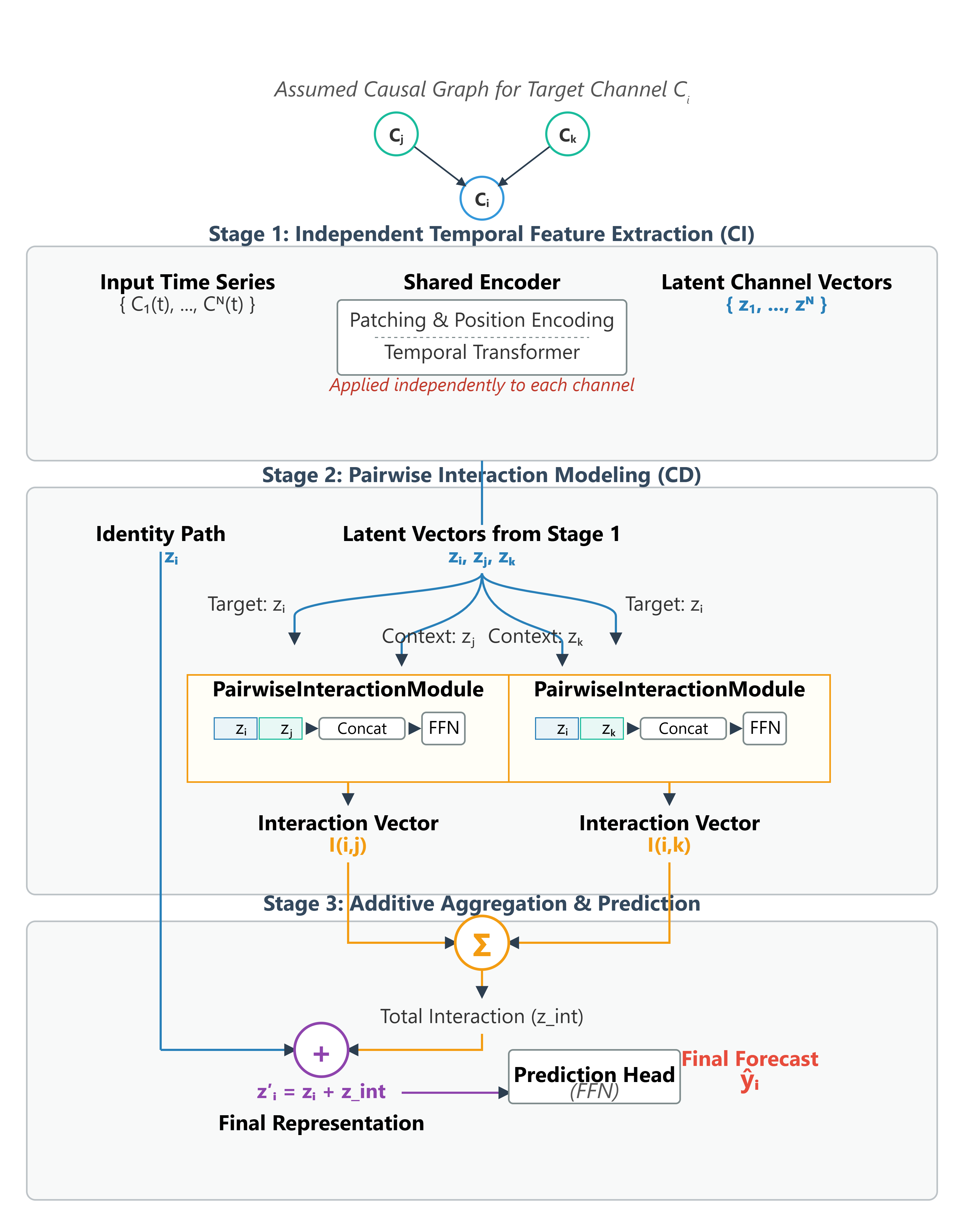}
    \caption{An overview of the proposed model architecture, which operates in three stages. 
\textbf{(1)} A CI module first extracts a latent vector $z_i$ for each input time series $C_i(t)$ using a shared temporal transformer. 
\textbf{(2)} Guided by a causal graph, pairwise mixing modules compute a vector $I(i,j)$ by mixing the latent representations of a cause node ($C_j$) and the target node ($C_i$). 
\textbf{(3)} Finally, these mixed vectors from all causal parents are aggregated into $z_\text{int}$ and added to the target's initial latent vector to form a final representation $z'_i$. This representation is then passed through a prediction head to generate the forecast $\hat{y}_i$.}
    \label{fig:cgpt_architecture}
\end{figure}

\noindent The model's task is to predict the future of a chosen target channel, $C_i$, based on its own history and the history of a pre-defined set of causal context channels, $\{C_j\}_{j \in \mathcal{J}}$. The input to the model $f_\theta$ is a sequence of past observations for these relevant channels over a \textit{context length} $L_{ctx}$. The model then outputs a forecast for the target channel, $\hat{y}_i$, over a given \textit{prediction horizon}. The target variable $C_i$ is typically a Key Performance Indicator (KPI), such as a critical process temperature or a downstream quality measurement. The context channels in $\mathcal{J}$ serve as covariates for the prediction task, providing the necessary information to model the system's dynamics accurately.

\subsection{Architectural Components}
\noindent The \modelname~model, denoted as $f_\theta$, operationalizes this predictive task in three distinct stages, as illustrated in Fig.~\ref{fig:cgpt_architecture}.

\noindent\textbf{1. Independent Temporal Feature Extraction (CI):}
Each input time series channel is first normalized individually using Reversible Instance Normalization (RevIN) \cite{Kim2022revin} within a context window; this dampens the impact of distributional shifts. The normalized series are then segmented into smaller non-overlapping patches, a technique inspired by PatchTST \cite{Nie2023patchtst} and commonly found in image processing domains. These patches serve as fundamental units for subsequent processing by the deep Transformer architecture. Specifically, an input time series $C_n \in \mathbb{R}^{L_{ctx}}$ is unfolded into a sequence of patches $\mat{C}_0^{(n)} \in \mathbb{R}^{N_p \times P}$, where $P$ is the patch size and $N_p$ is the number of patches. The initial token sequences for each channel are processed through a shared-weight Transformer encoder stack. Each channel's sequence is processed independently, meaning self-attention is computed only within each sequence. This design choice aligns with the core principle of Channel-Independent (CI) models \cite{Nie2023patchtst}, promoting the learning of generalizable temporal features. The encoder transforms the input tokens of each channel $C_n$ into a latent vector, denoted as $z_n$. These vectors encapsulate the temporal dynamics of each series, providing the basis for the subsequent interaction modeling.

\begin{equation}
\vect{z}_n = \text{Encoder}(\mat{C}_0^{(n)}) \quad \text{for } n \in \{1, \dots, N_{vars}\}
\end{equation}

\noindent\textbf{2. Pairwise Interaction Modeling (CD):}
This stage models how information from a set of causal parents, indexed by $\mathcal{J}$, is mixed with that of the target variable $C_i$. The core CD mechanism is a parameter-shared pairwise CD mixing module. It computes an interaction vector for each pair, $\vect{I}(i, j)$, by applying a feed-forward network (FFN) to the concatenation (denoted as $[\cdot, \cdot]$) of the target and parent latent vectors. \textbf{3. Additive Aggregation and Prediction:}
The individual interaction vectors, $\vect{I}(i, j)$, from all causal parents are aggregated via summation, denoted as $\vect{z}_\text{int}$ in Fig.~\ref{fig:cgpt_architecture}. This total interaction component is then additively combined with the target's original latent vector $\vect{z}_i$ to form the final representation, $\vect{z}'_i$. The entire process is captured in a single end-to-end model:
\begin{equation}
\label{eq:unified_model}
\hat{y}_i = \text{PredictionHead} \left( \vect{z}_i + \sum_{j \in \mathcal{J}} \vect{I}(i,j) \right)
\end{equation}
We use a simple summing operator to aggregate these individual mixed vectors. While this approach is straightforward and effective, more advanced concepts may be explored, such as learnable channel reconciliation strategies or more advanced mixing layers like pairwise cross-attention (see Sec.~\ref{sec:futurework}). \noindent Crucially, the entire model is trained end-to-end using a single loss function (e.g., MSE) on the final forecast $\hat{y}_i$. The decomposition is therefore learned implicitly through the architectural separation of information flows.

\section{Experiments}
\label{sec:experiments}
\noindent To validate our proposed pairwise interaction model, we designed a series of experiments to test its efficacy against common baselines and to understand the contributions of its core architectural components. 

\subsection{Experimental Setup}
\noindent We generate two synthetic multivariate time series datasets and also utilize four real-world datasets (see Tab.~\ref{tab:datasets}) from real-world industrial use-cases. The synthetic datasets include \textbf{Additive}, where the target channel is autoregressive, with its value determined by its own past state plus a linear combination of two of the three control channels and noise; the third control channel is included as a spurious correlate. The second dataset is \textbf{Interactive}, where the target is also autoregressive, determined by its own past state and a non-linear, multiplicative function of all three control channels, challenging linear summation assumptions. For the \textbf{real-world} scenarios, we utilize datasets from various industrial use-cases (see public \emph{ETTh1}~\cite{Zhou2021Informer}, \emph{Multi-Stage Factory}, $\copyright$~\emph{liveline.tech}, \emph{Amino Emissions}~\cite{jablonkamachine2023} and private \emph{Refial}, $\copyright$~\emph{grupo-otua.com}) and evaluate them on specific KPIs. The \emph{Refial} dataset, provided by our industrial partner \emph{Refial}, addresses the critical problem of optimizing an aluminum recycling tilting rotary furnace. This dataset, comprising 24500 time points with 5 channels and captures essential operational parameters. Specifically, it includes three  temperature measurements, the \textit{inside temperature} (our target), the \textit{hull temperature} and the \textit{exhaust temperature}, which are crucial for monitoring the thermal state. To control the kiln, the dataset provides two manipulable variables: \textit{gas flow}, which directly modulates heat input, and the binary \textit{door open/close} state, which influences heat retention. The scientific challenge lies in effectively modelling the complex and often non-linear causal relationships between these control variables and the resulting thermal dynamics to enhance energy efficiency and ensure optimal product quality in aluminum recycling.

\begin{table*}[h]
\centering
\begin{threeparttable}
\caption{Key characteristics of the selected datasets.}
\label{tab:dataset_statistics_compact}
\small
\begin{tabular}{@{}llcrcl@{}}
\toprule
\textbf{Dataset} & \textbf{Type} & \textbf{Channel} & \textbf{N} & \textbf{Target} & \textbf{\# Causes} \\
\midrule
ETTh1\tnote{1} ~\cite{Zhou2021Informer} & Public & 7 & 17420 & OT & --- \\
Kaggle Multi-Stage Fact.\tnote{2} & Public & 56 & 14088 & S1...0.U.Err. & --- \\
Carbon Amino Emissions\tnote{3} ~\cite{jablonkamachine2023} & Public & 67 & 5409 & 2-Amino...C4H11NO & --- \\
Refial\tnote{4} & Private & 5 & 24491 & Kiln Temp. & --- \\
\midrule
Synthetic Additive & Synthetic & 4 & 6144 & C3 & 2 (+1 distractor) \\
Synthetic Interactive & Synthetic & 4 & 6144 & C3 & 3 \\
\bottomrule
\end{tabular}
\begin{tablenotes}[para,flushleft]
  \footnotesize
  \tiny
  \item[1] \url{https://github.com/zhouhaoyi/ETDataset/blob/main/ETT-small/ETTh1.csv} \\
  \item[2] \url{https://www.kaggle.com/datasets/supergus/multistage-continuousflow-manufacturing-process} $\copyright$~\url{https://liveline.tech/}  \\
  \item[3] \url{https://github.com/kjappelbaum/aeml/blob/main/paper/20220210_smooth_window_16.pkl} \\
  \item[4] Private dataset from an EU H2020 project. \url{https://grupo-otua.com/} \\
\end{tablenotes}
\end{threeparttable}
\end{table*}
\label{tab:datasets}

\noindent We evaluate against two strong baselines (CI and CD) and three variants of \modelname. The baselines are \textbf{DLinear}, a univariate model (CI) that only uses the target's own history, and \textbf{MLP-Baseline}, a generic multivariate model (CD) that flattens all channel histories and processes them with a deep MLP. Our primary model, \textbf{LeakyPairwise \modelname~(Ours)}, combines CI and CD approaches by computing pairwise interactions using the target's own encoded history. We also test two ablations: \textbf{StrictPairwise \modelname}, where the target's history in the pairwise interaction module is replaced with a generic, learnable embedding for the interaction calculation, and \textbf{PureInfluence \modelname}, which predicts using only the sum of the resulting interaction vectors, making the model completely blind to the target's history in the final step.

\subsection{Training Environment and Hyperparameters}
\noindent All models are implemented in PyTorch \cite{PyTorch2019} and trained on a single NVIDIA A100 GPU. We use the AdamW optimizer \cite{AdamW2019} with a learning rate of $10^{-3}$ and a cosine annealing learning rate scheduler \cite{loshchilov2017sgdrstochasticgradientdescent}. We set for all reported models the patch size/stride to 32/32, i.e. non-overlapping patches, the model dimensions to 64 and the feed-forward networks to 128, use a single head and a single encoder layer. The networks are trained with a batch size of 256 for 100 epochs with an early stopping patience of 10 epochs and save the best model based on the validation set loss. The train, valid and test splits for the ETTh1 benchmark are mirrored from literature to enable comparison with state-of-the-art reported results, whereas for the other datasets, we use a train, valid and test split of 70\%,20\% and 10\%. For all experiments, we report MSE and MAE. 

\subsection{Experiment 1: Long-Term Forecasting with \modelname~vs. CD and CI Baselines}
\noindent For the long-term forecasting task (context=96, horizon=96), the results in Tab.~\ref{tab:experiment1} show that the \emph{CGPT} models, particularly \emph{LeakyPairwise} and \emph{StrictPairwise}, achieve state-of-the-art performance on most datasets. This suggests that for long-horizon prediction, their ability to capture complex system dynamics and inter-variable relationships provides a significant competitive advantage over simpler linear or unstructured CD models.

\begin{table}[ht!]
\centering
\scriptsize
\renewcommand{\arraystretch}{0.85}
\caption{Performance of forecasting models on the 96 context to 96 horizon task, averaged over 5 runs (Mean $\pm$ Std). Best results for each metric and dataset are in bold.}
\begin{tabularx}{\textwidth}{@{}llXXXX@{}}
\toprule
\multirow{2}{*}{\textbf{Dataset}} & \multirow{2}{*}{\textbf{Model}} & \multicolumn{2}{c}{\textbf{RevIN = No}} & \multicolumn{2}{c}{\textbf{RevIN = Yes}} \\
\cmidrule(lr){3-4} \cmidrule(lr){5-6}
& & \textbf{MAE} & \textbf{MSE} & \textbf{MAE} & \textbf{MSE} \\
\midrule
\multirow{5}{*}{Additive} & LeakyPairwise & 0.3165 $\pm$ 0.0001 & 0.1581 $\pm$ 0.0001 & 0.3188 $\pm$ 0.0016 & 0.1601 $\pm$ 0.0013 \\
 & StrictPairwise & \textbf{0.3164 $\pm$ 0.0001} & \textbf{0.1581 $\pm$ 0.0001} & 0.3182 $\pm$ 0.0013 & 0.1597 $\pm$ 0.0010 \\
 & PureInfluence & 0.3180 $\pm$ 0.0017 & 0.1593 $\pm$ 0.0014 & 0.3255 $\pm$ 0.0002 & 0.1671 $\pm$ 0.0001 \\
 & DLinear & 0.3181 $\pm$ 0.0001 & 0.1598 $\pm$ 0.0001 & 0.3180 $\pm$ 0.0000 & 0.1598 $\pm$ 0.0001 \\
 & MLP & 0.3177 $\pm$ 0.0002 & 0.1592 $\pm$ 0.0001 & 0.3251 $\pm$ 0.0004 & 0.1668 $\pm$ 0.0003 \\
\midrule
\multirow{5}{*}{Carbon} & LeakyPairwise & \textbf{0.2711 $\pm$ 0.0162} & \textbf{0.1344 $\pm$ 0.0055} & 0.3522 $\pm$ 0.0213 & 0.2087 $\pm$ 0.0190 \\
 & StrictPairwise & 0.2988 $\pm$ 0.0286 & 0.1570 $\pm$ 0.0252 & 0.3051 $\pm$ 0.0068 & 0.1754 $\pm$ 0.0084 \\
 & PureInfluence & 0.3362 $\pm$ 0.0647 & 0.1894 $\pm$ 0.0478 & 0.3554 $\pm$ 0.0172 & 0.2173 $\pm$ 0.0103 \\
 & DLinear & 0.2948 $\pm$ 0.0016 & 0.1550 $\pm$ 0.0009 & 0.3184 $\pm$ 0.0011 & 0.1883 $\pm$ 0.0009 \\
 & MLP & 0.3208 $\pm$ 0.0234 & 0.1641 $\pm$ 0.0136 & 0.3650 $\pm$ 0.0065 & 0.2466 $\pm$ 0.0119 \\
\midrule
\multirow{5}{*}{ETTh1} & LeakyPairwise & 0.5611 $\pm$ 0.1009 & 0.4122 $\pm$ 0.1300 & 0.1843 $\pm$ 0.0021 & 0.0575 $\pm$ 0.0007 \\
 & StrictPairwise & 0.3819 $\pm$ 0.0289 & 0.2036 $\pm$ 0.0263 & 0.1847 $\pm$ 0.0014 & 0.0580 $\pm$ 0.0009 \\
 & PureInfluence & 1.5919 $\pm$ 0.1403 & 3.2648 $\pm$ 0.4179 & 0.1982 $\pm$ 0.0012 & 0.0662 $\pm$ 0.0003 \\
 & DLinear & 0.1863 $\pm$ 0.0047 & 0.0633 $\pm$ 0.0034 & \textbf{0.1782 $\pm$ 0.0003} & \textbf{0.0556 $\pm$ 0.0002} \\
 & MLP & 0.8062 $\pm$ 0.0854 & 0.7473 $\pm$ 0.1433 & 0.1873 $\pm$ 0.0016 & 0.0601 $\pm$ 0.0011 \\
\midrule
\multirow{5}{*}{Interactive} & LeakyPairwise & \textbf{0.8992 $\pm$ 0.0004} & \textbf{1.8751 $\pm$ 0.0006} & 0.9064 $\pm$ 0.0024 & 1.8935 $\pm$ 0.0075 \\
 & StrictPairwise & 0.8993 $\pm$ 0.0005 & 1.8753 $\pm$ 0.0009 & 0.9075 $\pm$ 0.0019 & 1.8932 $\pm$ 0.0053 \\
 & PureInfluence & 0.8996 $\pm$ 0.0002 & 1.8898 $\pm$ 0.0005 & 0.9482 $\pm$ 0.0003 & 1.9804 $\pm$ 0.0009 \\
 & DLinear & 0.9129 $\pm$ 0.0005 & 1.8965 $\pm$ 0.0013 & 0.9145 $\pm$ 0.0003 & 1.8998 $\pm$ 0.0006 \\
 & MLP & 0.9000 $\pm$ 0.0004 & 1.8807 $\pm$ 0.0008 & 0.9372 $\pm$ 0.0008 & 1.9486 $\pm$ 0.0028 \\
\midrule
\multirow{5}{*}{Kaggle} & LeakyPairwise & 0.0962 $\pm$ 0.0156 & 0.0153 $\pm$ 0.0056 & 0.0327 $\pm$ 0.0002 & 0.0022 $\pm$ 0.0000 \\
 & StrictPairwise & 0.1652 $\pm$ 0.0106 & 0.0410 $\pm$ 0.0054 & 0.0334 $\pm$ 0.0005 & 0.0022 $\pm$ 0.0000 \\
 & PureInfluence & 0.1654 $\pm$ 0.0066 & 0.0410 $\pm$ 0.0039 & 0.0331 $\pm$ 0.0009 & 0.0022 $\pm$ 0.0001 \\
 & DLinear & 0.0622 $\pm$ 0.0016 & 0.0057 $\pm$ 0.0002 & \textbf{0.0319 $\pm$ 0.0000} & \textbf{0.0021 $\pm$ 0.0000} \\
 & MLP & 1.1098 $\pm$ 0.3495 & 1.5773 $\pm$ 0.8458 & 0.0325 $\pm$ 0.0001 & 0.0022 $\pm$ 0.0000 \\
\midrule
\multirow{5}{*}{Refial} & LeakyPairwise & \textbf{0.1652 $\pm$ 0.0067} & 0.0756 $\pm$ 0.0047 & 0.1674 $\pm$ 0.0053 & 0.0982 $\pm$ 0.0027 \\
 & StrictPairwise & 0.1714 $\pm$ 0.0062 & 0.0788 $\pm$ 0.0041 & 0.1730 $\pm$ 0.0034 & 0.1019 $\pm$ 0.0032 \\
 & PureInfluence & 0.2238 $\pm$ 0.0080 & 0.0958 $\pm$ 0.0062 & 0.2239 $\pm$ 0.0063 & 0.1384 $\pm$ 0.0029 \\
 & DLinear & 0.2698 $\pm$ 0.0027 & 0.1692 $\pm$ 0.0016 & 0.2767 $\pm$ 0.0013 & 0.1897 $\pm$ 0.0003 \\
 & MLP & 0.1699 $\pm$ 0.0067 & \textbf{0.0721 $\pm$ 0.0028} & 0.1795 $\pm$ 0.0049 & 0.1047 $\pm$ 0.0037 \\
\bottomrule
\end{tabularx}
\label{tab:experiment1}
\end{table}

\newpage

\subsection{Experiment 2: One-Step Forecasting with \modelname~vs CD and CI Baselines}
\noindent For the one-step-ahead forecasting task (context=96, horizon=1), the results in Tab.~\ref{tab:experiment2} show a clear preference for simplicity. The DLinear model, which excels at capturing autoregressive patterns, decisively outperforms the more complex CD models on most datasets. Among the remaining models, the \emph{LeakyPairwise} and \emph{StrictPairwise} variants also show a clear advantage over the standard MLP, performing better on most datasets.

\begin{table}[h!]
\centering
\scriptsize
\renewcommand{\arraystretch}{0.85}
\caption{Performance of forecasting models on the 96 context to 1 horizon task, averaged over 5 runs (Mean $\pm$ Std). Best results for each metric and dataset are in bold.}
\begin{tabularx}{\textwidth}{@{}llXXXX@{}}
\toprule
\multirow{2}{*}{\textbf{Dataset}} & \multirow{2}{*}{\textbf{Model}} & \multicolumn{2}{c}{\textbf{RevIN = No}} & \multicolumn{2}{c}{\textbf{RevIN = Yes}} \\
\cmidrule(lr){3-4} \cmidrule(lr){5-6}
& & \textbf{MAE} & \textbf{MSE} & \textbf{MAE} & \textbf{MSE} \\
\midrule
\multirow{5}{*}{Additive} & LeakyPairwise & 0.0686 $\pm$ 0.0037 & 0.0073 $\pm$ 0.0008 & 0.0699 $\pm$ 0.0012 & 0.0077 $\pm$ 0.0002 \\
 & StrictPairwise & 0.0697 $\pm$ 0.0043 & 0.0075 $\pm$ 0.0009 & 0.0692 $\pm$ 0.0011 & 0.0075 $\pm$ 0.0001 \\
 & PureInfluence & 0.1032 $\pm$ 0.0013 & 0.0168 $\pm$ 0.0003 & 0.1099 $\pm$ 0.0004 & 0.0201 $\pm$ 0.0003 \\
 & DLinear & 0.0875 $\pm$ 0.0013 & 0.0118 $\pm$ 0.0003 & 0.0879 $\pm$ 0.0013 & 0.0119 $\pm$ 0.0003 \\
 & MLP & \textbf{0.0616 $\pm$ 0.0012} & \textbf{0.0060 $\pm$ 0.0002} & 0.0748 $\pm$ 0.0008 & 0.0088 $\pm$ 0.0002 \\
\midrule
\multirow{5}{*}{Carbon} & LeakyPairwise & 0.1110 $\pm$ 0.0115 & 0.0533 $\pm$ 0.0041 & 0.1065 $\pm$ 0.0055 & 0.0512 $\pm$ 0.0024 \\
 & StrictPairwise & 0.1368 $\pm$ 0.0117 & 0.0663 $\pm$ 0.0106 & 0.1121 $\pm$ 0.0078 & 0.0525 $\pm$ 0.0031 \\
 & PureInfluence & 0.4028 $\pm$ 0.0192 & 0.2854 $\pm$ 0.0144 & 0.3137 $\pm$ 0.0112 & 0.1728 $\pm$ 0.0034 \\
 & DLinear & 0.0982 $\pm$ 0.0032 & 0.0487 $\pm$ 0.0023 & \textbf{0.0960 $\pm$ 0.0016} & \textbf{0.0483 $\pm$ 0.0014} \\
 & MLP & 0.3954 $\pm$ 0.0417 & 0.2738 $\pm$ 0.0304 & 0.2864 $\pm$ 0.0127 & 0.1409 $\pm$ 0.0058 \\
\midrule
\multirow{5}{*}{ETTh1} & LeakyPairwise & 0.0736 $\pm$ 0.0081 & 0.0091 $\pm$ 0.0016 & 0.0472 $\pm$ 0.0003 & 0.0041 $\pm$ 0.0001 \\
 & StrictPairwise & 0.0736 $\pm$ 0.0083 & 0.0089 $\pm$ 0.0016 & 0.0472 $\pm$ 0.0001 & 0.0041 $\pm$ 0.0000 \\
 & PureInfluence & 1.7328 $\pm$ 0.1435 & 3.5726 $\pm$ 0.5191 & 0.1519 $\pm$ 0.0025 & 0.0372 $\pm$ 0.0011 \\
 & DLinear & 0.0459 $\pm$ 0.0001 & 0.0039 $\pm$ 0.0000 & \textbf{0.0457 $\pm$ 0.0000} & \textbf{0.0039 $\pm$ 0.0000} \\
 & MLP & 0.2526 $\pm$ 0.0362 & 0.0897 $\pm$ 0.0230 & 0.0617 $\pm$ 0.0014 & 0.0065 $\pm$ 0.0003 \\
\midrule
\multirow{5}{*}{Interactive} & LeakyPairwise & 0.3435 $\pm$ 0.0062 & 0.2344 $\pm$ 0.0084 & 0.3415 $\pm$ 0.0032 & 0.2323 $\pm$ 0.0053 \\
 & StrictPairwise & 0.3462 $\pm$ 0.0067 & 0.2381 $\pm$ 0.0068 & 0.3465 $\pm$ 0.0056 & 0.2357 $\pm$ 0.0044 \\
 & PureInfluence & 0.7193 $\pm$ 0.0010 & 1.1550 $\pm$ 0.0012 & 0.7475 $\pm$ 0.0015 & 1.1731 $\pm$ 0.0028 \\
 & DLinear & 0.3231 $\pm$ 0.0010 & 0.2133 $\pm$ 0.0007 & \textbf{0.3229 $\pm$ 0.0010} & \textbf{0.2131 $\pm$ 0.0007} \\
 & MLP & 0.3716 $\pm$ 0.0041 & 0.2661 $\pm$ 0.0019 & 0.4044 $\pm$ 0.0058 & 0.3112 $\pm$ 0.0068 \\
\midrule
\multirow{5}{*}{Kaggle} & LeakyPairwise & 0.0446 $\pm$ 0.0105 & 0.0046 $\pm$ 0.0014 & 0.0324 $\pm$ 0.0013 & 0.0034 $\pm$ 0.0004 \\
 & StrictPairwise & 0.0919 $\pm$ 0.0503 & 0.0133 $\pm$ 0.0097 & 0.0319 $\pm$ 0.0010 & 0.0034 $\pm$ 0.0003 \\
 & PureInfluence & 0.1731 $\pm$ 0.0854 & 0.0396 $\pm$ 0.0313 & 0.0359 $\pm$ 0.0003 & 0.0044 $\pm$ 0.0001 \\
 & DLinear & 0.0308 $\pm$ 0.0010 & 0.0023 $\pm$ 0.0001 & \textbf{0.0253 $\pm$ 0.0001} & \textbf{0.0018 $\pm$ 0.0000} \\
 & MLP & 0.2285 $\pm$ 0.1027 & 0.0698 $\pm$ 0.0570 & 0.0365 $\pm$ 0.0003 & 0.0044 $\pm$ 0.0001 \\
\midrule
\multirow{5}{*}{Refial} & LeakyPairwise & 0.0055 $\pm$ 0.0005 & 0.0001 $\pm$ 0.0000 & 0.0043 $\pm$ 0.0005 & 0.0000 $\pm$ 0.0000 \\
 & StrictPairwise & 0.0061 $\pm$ 0.0010 & 0.0001 $\pm$ 0.0000 & 0.0040 $\pm$ 0.0009 & \textbf{0.0000 $\pm$ 0.0000} \\
 & PureInfluence & 0.1875 $\pm$ 0.0309 & 0.0528 $\pm$ 0.0142 & 0.0442 $\pm$ 0.0030 & 0.0041 $\pm$ 0.0002 \\
 & DLinear & 0.0046 $\pm$ 0.0002 & 0.0000 $\pm$ 0.0000 & \textbf{0.0038 $\pm$ 0.0001} & 0.0000 $\pm$ 0.0000 \\
 & MLP & 0.0186 $\pm$ 0.0045 & 0.0007 $\pm$ 0.0003 & 0.0131 $\pm$ 0.0007 & 0.0004 $\pm$ 0.0000 \\
\bottomrule
\end{tabularx}
\label{tab:experiment2}
\end{table}

\subsection{Experiment 3: Impact of Subsequent Removal of Gradient Flows Involving Target History}

\noindent In this experiment (see Tab.~\ref{tab:experimentdegardation}), we evaluate how much \modelname~degrades by removing subsequently architectural components that allow gradient flow w.r.t the target history. The results show that completely removing the model's access to the target's past values, i.e. \emph{PureInfluence}, causes a significant degradation in performance across all datasets and forecasting horizons. This confirms that the autoregressive signal is a  critical feature for forecasting, and that treating the problem as a purely extrinsic regression is ineffective. In contrast, the performance difference between the full model (\emph{LeakyPairwise}) and the version with restricted internal gradient flow (\emph{StrictPairwise}) is consistently marginal, suggesting this specific pathway is a minor refinement rather than a core performance driver.

\begin{sidewaystable}[!ht]
\centering
\scriptsize 
\setlength{\tabcolsep}{2pt} 
\caption{Performance degradation of pairwise model variants across two forecasting tasks (96-96 and 96-1). We compare LeakyPairwise (\modelname) against versions with restricted gradient flow: StrictPairwise (no intra-module flow) and PureInfluence (no target history connection). PureInfluence \modelname~in this experiment is equivalent to an extrinsic regression problem. Results are averaged over 5 runs (Mean $\pm$ Std). Best results among the three variants for each metric are in bold.}
\label{tab:experiment3_updated}
\begin{tabular}{@{}llcccc|cccc@{}}
\toprule
& & \multicolumn{4}{c}{\textbf{Experiment 1: 96 $\rightarrow$ 96}} & \multicolumn{4}{c}{\textbf{Experiment 2: 96 $\rightarrow$ 1}} \\
\cmidrule(lr){3-6} \cmidrule(lr){7-10}
& & \multicolumn{2}{c}{\textbf{RevIN = No}} & \multicolumn{2}{c}{\textbf{RevIN = Yes}} & \multicolumn{2}{c}{\textbf{RevIN = No}} & \multicolumn{2}{c}{\textbf{RevIN = Yes}} \\
\cmidrule(lr){3-4} \cmidrule(lr){5-6} \cmidrule(lr){7-8} \cmidrule(lr){9-10}
\textbf{Dataset} & \textbf{Model} & \textbf{MAE} & \textbf{MSE} & \textbf{MAE} & \textbf{MSE} & \textbf{MAE} & \textbf{MSE} & \textbf{MAE} & \textbf{MSE} \\
\midrule
\multirow{3}{*}{Additive} & LeakyPairwise & 0.316 $\pm$ 0.000 & 0.158 $\pm$ 0.000 & 0.319 $\pm$ 0.002 & 0.160 $\pm$ 0.001 & \textbf{0.069 $\pm$ 0.004} & \textbf{0.007 $\pm$ 0.001} & 0.070 $\pm$ 0.001 & 0.008 $\pm$ 0.000 \\
 & StrictPairwise & \textbf{0.316 $\pm$ 0.000} & \textbf{0.158 $\pm$ 0.000} & \textbf{0.318 $\pm$ 0.001} & \textbf{0.160 $\pm$ 0.001} & 0.070 $\pm$ 0.004 & 0.007 $\pm$ 0.001 & \textbf{0.069 $\pm$ 0.001} & \textbf{0.008 $\pm$ 0.000} \\
 & PureInfluence & 0.318 $\pm$ 0.002 & 0.159 $\pm$ 0.001 & 0.326 $\pm$ 0.000 & 0.167 $\pm$ 0.000 & 0.103 $\pm$ 0.001 & 0.017 $\pm$ 0.000 & 0.110 $\pm$ 0.000 & 0.020 $\pm$ 0.000 \\
\midrule
\multirow{3}{*}{Carbon} & LeakyPairwise & \textbf{0.271 $\pm$ 0.016} & \textbf{0.134 $\pm$ 0.006} & 0.352 $\pm$ 0.021 & 0.209 $\pm$ 0.019 & \textbf{0.111 $\pm$ 0.012} & \textbf{0.053 $\pm$ 0.004} & \textbf{0.107 $\pm$ 0.005} & \textbf{0.051 $\pm$ 0.002} \\
 & StrictPairwise & 0.299 $\pm$ 0.029 & 0.157 $\pm$ 0.025 & \textbf{0.305 $\pm$ 0.007} & \textbf{0.175 $\pm$ 0.008} & 0.137 $\pm$ 0.012 & 0.066 $\pm$ 0.011 & 0.112 $\pm$ 0.008 & 0.053 $\pm$ 0.003 \\
 & PureInfluence & 0.336 $\pm$ 0.065 & 0.189 $\pm$ 0.048 & 0.355 $\pm$ 0.017 & 0.217 $\pm$ 0.010 & 0.403 $\pm$ 0.019 & 0.285 $\pm$ 0.014 & 0.314 $\pm$ 0.011 & 0.173 $\pm$ 0.003 \\
\midrule
\multirow{3}{*}{ETTh1} & LeakyPairwise & 0.561 $\pm$ 0.101 & 0.412 $\pm$ 0.130 & \textbf{0.184 $\pm$ 0.002} & \textbf{0.057 $\pm$ 0.001} & 0.074 $\pm$ 0.008 & 0.009 $\pm$ 0.002 & \textbf{0.047 $\pm$ 0.000} & 0.004 $\pm$ 0.000 \\
 & StrictPairwise & \textbf{0.382 $\pm$ 0.029} & \textbf{0.204 $\pm$ 0.026} & 0.185 $\pm$ 0.001 & 0.058 $\pm$ 0.001 & \textbf{0.074 $\pm$ 0.008} & \textbf{0.009 $\pm$ 0.002} & 0.047 $\pm$ 0.000 & \textbf{0.004 $\pm$ 0.000} \\
 & PureInfluence & 1.592 $\pm$ 0.140 & 3.265 $\pm$ 0.418 & 0.198 $\pm$ 0.001 & 0.066 $\pm$ 0.000 & 1.733 $\pm$ 0.144 & 3.573 $\pm$ 0.519 & 0.152 $\pm$ 0.002 & 0.037 $\pm$ 0.001 \\
\midrule
\multirow{3}{*}{Interactive} & LeakyPairwise & \textbf{0.899 $\pm$ 0.000} & \textbf{1.875 $\pm$ 0.001} & \textbf{0.906 $\pm$ 0.002} & 1.894 $\pm$ 0.008 & \textbf{0.344 $\pm$ 0.006} & \textbf{0.234 $\pm$ 0.008} & \textbf{0.341 $\pm$ 0.003} & \textbf{0.232 $\pm$ 0.005} \\
 & StrictPairwise & 0.899 $\pm$ 0.001 & 1.875 $\pm$ 0.001 & 0.907 $\pm$ 0.002 & \textbf{1.893 $\pm$ 0.005} & 0.346 $\pm$ 0.007 & 0.238 $\pm$ 0.007 & 0.346 $\pm$ 0.006 & 0.236 $\pm$ 0.004 \\
 & PureInfluence & 0.900 $\pm$ 0.000 & 1.890 $\pm$ 0.000 & 0.948 $\pm$ 0.000 & 1.980 $\pm$ 0.001 & 0.719 $\pm$ 0.001 & 1.155 $\pm$ 0.001 & 0.748 $\pm$ 0.002 & 1.173 $\pm$ 0.003 \\
\midrule
\multirow{3}{*}{Kaggle} & LeakyPairwise & \textbf{0.096 $\pm$ 0.016} & \textbf{0.015 $\pm$ 0.006} & \textbf{0.033 $\pm$ 0.000} & \textbf{0.002 $\pm$ 0.000} & \textbf{0.045 $\pm$ 0.010} & \textbf{0.005 $\pm$ 0.001} & 0.032 $\pm$ 0.001 & 0.003 $\pm$ 0.000 \\
 & StrictPairwise & 0.165 $\pm$ 0.011 & 0.041 $\pm$ 0.005 & 0.033 $\pm$ 0.000 & 0.002 $\pm$ 0.000 & 0.092 $\pm$ 0.050 & 0.013 $\pm$ 0.010 & \textbf{0.032 $\pm$ 0.001} & \textbf{0.003 $\pm$ 0.000} \\
 & PureInfluence & 0.165 $\pm$ 0.007 & 0.041 $\pm$ 0.004 & 0.033 $\pm$ 0.001 & 0.002 $\pm$ 0.000 & 0.173 $\pm$ 0.085 & 0.040 $\pm$ 0.031 & 0.036 $\pm$ 0.000 & 0.004 $\pm$ 0.000 \\
\midrule
\multirow{3}{*}{Refial} & LeakyPairwise & \textbf{0.1652 $\pm$ 0.0067} & \textbf{0.0756 $\pm$ 0.0047} & \textbf{0.1674 $\pm$ 0.0053} & \textbf{0.0982 $\pm$ 0.0027} & \textbf{0.0055 $\pm$ 0.0005} & \textbf{0.0001 $\pm$ 0.0000} & 0.0043 $\pm$ 0.0005 & 0.0000 $\pm$ 0.0000 \\
 & StrictPairwise & 0.1714 $\pm$ 0.0062 & 0.0788 $\pm$ 0.0041 & 0.1730 $\pm$ 0.0034 & 0.1019 $\pm$ 0.0032 & 0.0061 $\pm$ 0.0010 & 0.0001 $\pm$ 0.0000 & \textbf{0.0040 $\pm$ 0.0009} & \textbf{0.0000 $\pm$ 0.0000} \\
 & PureInfluence & 0.2238 $\pm$ 0.0080 & 0.0958 $\pm$ 0.0062 & 0.2239 $\pm$ 0.0063 & 0.1384 $\pm$ 0.0029 & 0.1875 $\pm$ 0.0309 & 0.0528 $\pm$ 0.0142 & 0.0442 $\pm$ 0.0030 & 0.0041 $\pm$ 0.0002 \\
\bottomrule
\end{tabular}
\label{tab:experimentdegardation}
\end{sidewaystable}

\FloatBarrier

\section{Discussion and Future Work}
\label{sec:futurework}
Our experiments reveal that for one-step-ahead prediction, the results show that a simple channel-independent model like DLinear is superior. This suggests that for predicting the immediate future, the dominant autoregressive signal in the target's history is paramount, and the overhead of modeling complex cross-channel interactions can be a detrimental distraction. In the more challenging long-term forecasting task, this univariate reliance is insufficient and our proposed \modelname~architecture excels by effectively modeling the complex, dynamic relationships between the target and its causal drivers, outperforming both tested CI and CD baselines. 
\noindent A key target for future work is to scale \modelname~into a true industrial foundation model by pre-training on vast and diverse cross-domain industrial datasets. This effort will be centred on promising research directions including (i) \textbf{self-supervised learning}, which introduces e.g. contrastive supervisory signals during pre-training to pull representations of e.g. causally-linked variables closer in the latent space than non-causal pairs, (ii) \textbf{probabilistic modelling} for uncertainty quantification, and (iii) \textbf{learnable influence aggregation}, which replaces the current summation of causal influences with an attention-based mechanism that can weight context variables based on the target's state. Additionally we plan to explore more sophisticated methods for the interaction function, e.g. \textbf{patch-based cross-attention}.
\noindent Beyond this core vision, other research directions include a hierarchical metadata schema to enhance generalization during pre-training. The pre-training phase may leverage high-level, low-cardinality semantic concepts (e.g., Variable Type, Causal Role) to learn universal process dynamics of metadata commonly shared across industrial use-cases. This may also include high-level process-state data or product-state data. Subsequently, the model will be specialized for downstream tasks via parameter-efficient fine-tuning, incorporating fine-grained, high-cardinality metadata (e.g., Variable ID, Product Grade) using techniques like adapter layers.

\section{Conclusion}
\label{sec:discussion}
\noindent We introduced Causally-Guided Pairwise Transformer (\modelname), a novel architecture for modelling industrial Digital Twin use-cases that effectively combines the principles of channel-independent processing with channel-dependent interaction modeling. Its key innovation is an "any-variate" architecture that processes inputs as variable pairs while producing a fixed, univariate output—a design specifically tailored for common industrial tasks like Key Performance Indicator (KPI) prediction. This makes the architecture agnostic to the number of input sensors—a critical requirement for diverse industrial applications—and overcomes the limitations of fixed-input models like a standard MLP. Experiments demonstrate that \modelname~excels at long-term forecasting by leveraging causal drivers, outperforming both channel-independent and channel-dependent baselines. By handling arbitrary sensor configurations without architectural changes, \modelname~represents a significant step towards a versatile, "one-for-all" predictive model for the process industry.
\noindent The prerequisite of our framework is the existence of a causal graph. For datasets without a given causal graph, all variables are assumed to be causal drivers w.r.t. the modelled target. If no graph is given in a multivariate to multivariate modelling problem, the number of pairs results in a combinatorial explosion and becomes computationally infeasible very fast. Thus, we focus on multivariate to single variate problems, a use-case commonly encountered in industrial systems, and computationally feasible without a known causal graph. 

\section{Acknowledgments}
\noindent This work received funding as part of the Trineflex project (trineflex.eu), which has received funding from the European Union’s Horizon Europe research and innovation programme under Grant Agreement No 101058174. Funded by the European Union. In addition this work was supported by the Austrian ministries BMIMI, BMWET and the State of Upper Austria in the frame of the SCCH COMET competence center INTEGRATE (FFG 892418).

\bibliographystyle{elsarticle-num} 
\bibliography{references} 
\end{document}